# Improving Road Signs Detection performance by Combining the Features of Hough Transform and Texture

Ayaou Tarik
Department of physics
Faculty of Sciences, Ibn Zohr University
Agadir, Morocco
tarikayaou@gmail.com

Mourad Boussaid
Department of Computer Science
Faculty of Sciences, Ibn Zohr University
Agadir, Morocco

Afdel Karim
Department of Computer Science
Faculty of Sciences, Ibn Zohr University
Agadir, Morocco
kafdel@ymail.com

Amghar Abdellah
Department of physics
Faculty of Sciences, Ibn Zohr University
Agadir, Morocco
a.amghar@uiz.ac.ma

*Abstract*—With the large uses of the intelligent systems in different domains, and in order to increase the drivers and pedestrians' safety, the road and traffic sign recognition system has been a challenging issue and an important task for many years. But studies, done in this field of detection and recognition of traffic signs in an image, which are interested in the Arab context, are still insufficient. Detection of the road signs present in the scene is the one of the main stages of the traffic sign detection and recognition. In this paper, an efficient solution to enhance road signs detection, including Arabic context, performance based on color segmentation, Randomized Hough Transform and the combination of Zernike moments and Haralick features has been made. Segmentation stage is useful to determine the Region of Interest (ROI) in the image. The Randomized Hough Transform (RHT) is used to detect the circular and octagonal shapes. This stage is improved by the extraction of the Haralick features and Zernike moments. Furthermore, we use it as input of a classifier based on SVM. Experimental results show that the proposed approach allows us to perform the measurement's precision.

## I- Introduction

A large number of deaths and injuries are caused by road accidents. So as to reduce the number of accidents in the road, in recent years, road sign recognition has known considerable scientists attention due to its importance as an intelligent system. For this reason, the development of a robust monitoring system, adapted to the Arab context, to help drivers remains an essential task. There are many traffic signs detection and recognition systems which aim to monitor the road. However, the detection stage still needs more improvements. This stage represents an essential task in order to develop an Intelligent Transport System (ITS). Moreover, the visibility of traffic signs is very important for the safety of drivers and pedestrians.

Generally, there are three main stages to identify traffic signs in the image: segmentation, detection and recognition. The first stage is useful to reduce the search space and isolate the Region of Interest (ROI), although the detection block [1] selects those blobs that have an appropriate traffic sign shape and the recognition stage identifies the category of the pictogram already detected. Many research focused on improving the recognition stage, although the main one is to improve the detection block because the quality of the detection is necessary to get a good recognition [2].

To detect possible sign candidates and recognized it, there are often three categories of processing:

- Image segmentation: Detection of regions of interest based on color [3] [4] [5] [6].
- Geometry Methods: The analysis of image contours using mathematical transforms (Hough Transform, Harris-Stephens method …) [3] [4] [7].
- Learning Methods: The training of a classifier (SVM, MLP, KNN …) on a database [8] [9].

In the identification of signs field, some strategies use segmentation based on Hue-Saturation –Value (HSV) space [3] [10] because it is considered largely invariant to illumination changes. Also, as HSV, RGB normalized used [11]. Furthermore, using the Hough transform, as a robust geometric method for detecting circles (speed limit signs or stop) or lines (triangular panels) [12] is useful. The corner detection is also employed [7]. To recognize a road sign, many approaches use 2D features-based methods, such as Zernike moments [13], Fast Fourier Transform (FFT) [14] and Local Energy based Shape Histogram (LESH) [15]. Likewise, the learning-based methods are used, in both detection and recognition [8] [9] [16].In this study, two phases are applied to detect signs. The first is to reduce the search space in the image using a robust segmentation, and the second aims to detect the shapes (circular and octagonal) in regions of interest using RHT. However, this technique generates much false detection. For this reason we use the Haralick features and Zernike moments, which are invariant to rotation, as input of an SVM classifier to perform the precision of road signs detection.

In the Following sections, details of our detection method are described. Then, Section 3 reports experimental results and discuss it. Finally, a conclusion is presented in Section 4.

## II- System overview

In this paper, we describe our work on traffic sign detection. The proposed system allows us to effectively detect road signs from a video sequence. It consists of two main stages:
- Segmentation
- Detection

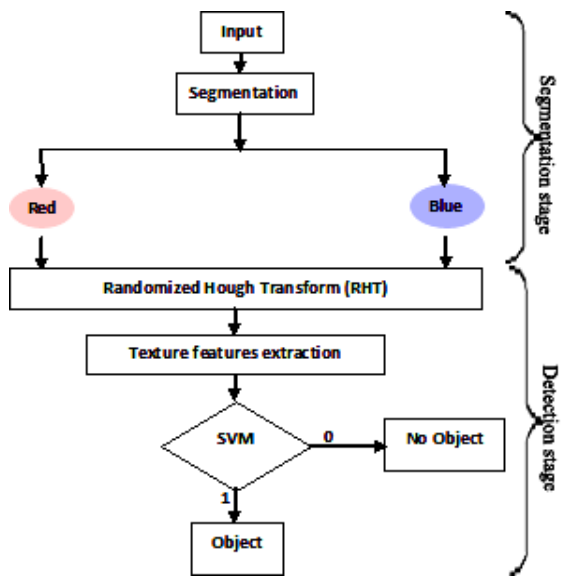

Figure1. The general strategy adopted for this work.

### 1- Segmentation

The traffic signs are often colored in red, blue or yellow. Then, the segmentation stage is crucial to isolate the Region of Interest. In this stage, the exploitation of color information as a feature is common implementation.

Frequently, the main difficulties faced by researchers are due to illumination changes which are caused by the color of daylight change depending on whether conditions [17], e.g., sky with/without clouds, time of the day, and night when all sort of artificial lights are surrounded. Since the direct thresholding of the RGB channels is sensitive to changes in illumination, the relation between the RGB (Red Green Blue) colors is often used [8]. The RGB-HSV is very useful, but its conversion formulas are non-linear and hence the computational cost involved is too high. We choose to implement the color enhancement method proposed by Ruta et al. [9] and to use also the morphological filters.

For each RGB pixel $x = [X_R, X_G, X_B]$ and $S = X_R + X_G + X_B$ a simple color enhancement is provided by:
- Red color enhancement: $f_R(X) = \max(0, \min(X_R - X_G, X_R - X_B)/S)$
- Blue color enhancement: $f_B(X) = \max(0, \min(X_B - X_G, X_B - X_G)/S)$ (1)

This technique is very adequate for our case as a fast, robust and simple one. It avoids expensive computation of the whole gradient magnitude map which, with the exception of the sparse edge pixels, is of no use to the shape detector [9]. Furthermore, we applied an adaptive global threshold, based on the mean and standard deviation value, to the enhanced image which result a binary mask of the ROIs. The global mean helps take in the account the illumination of the whole image, while a mean computed locally is more sensitive to small illumination variations in the image [8].

An example of image segmentation is shown in the figure 2; it illustrates the robustness of this method applied to the RGB traffic images.

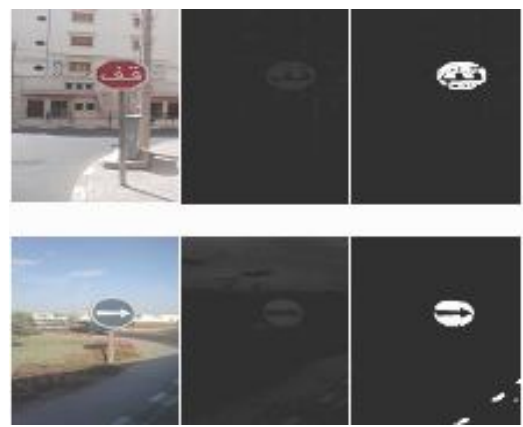

Figure2. The filtered images (Middle), using the transforms defined in (1), of the original images (left). In the right, the binaries images are shown.

### 2- Detection

The detection stage can be divided into two levels. The blob detection level using Randomized Hough Transform (RHT) is used to detect circular and octagonal shapes present in the image. Furthermore, the object classification level is used to extend the detection performance by using SVM classifier with Haralick and Zernike features. So as to reduce the feature space dimension, we use the Principal Component Analysis (PCA).

### a- Randomized Hough Transform

The classical Hough Transform (SHT) represents a robust method to detect regular shapes. Although, it is computationally complex, it is unsuitable for many real time applications.

Lie Xu et al. [18] proved that RHT, in comparison with HT, has high speed and small storage. In this work, we focus on the detection of circular and octagonal road signs. The octagonal signs can be considered as regular ellipse with the finite sides. The performance of RHT is increased by applying the image segmentation intended to reduce the space research and isolate the regions of interest and further [9] enhance the edges of specific color within each.

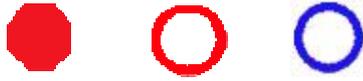

Figure3. Road signs criteria in this paper

Generally, in the RHT, three pixels $A(x_1, y_1)$, $B(x_2, y_2)$, and $C(x_3, y_3)$, as shown in figure below, are randomly chosen from the image and further estimate the tangent of each pixel.

The characteristic equation of an ellipse is written as:

$$a(x-p)^2 + 2b(x-p)(y-p) + c(y-p)^2 = 1 \quad (2)$$

In order to compute the parameters of the ellipse, we must find the center $O(p, q)$, and the three parameters a, b and c with the restriction that: $ac - b^2 > 0$.

To find the center of the ellipse, the method of Yuen et al. [19] is used; The point of intersection of the two lines nm and st represents the center O of the ellipse.

With:

n : the intersection of the tangent at the point C and the tangent at the point B.

m: the midpoint of the line segment from B to C.

s: the intersection of the tangent at the point A and the tangent at the point B.

t: the midpoint of the line segment from A to B.

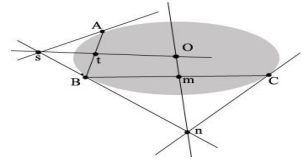

Figure4. Using three points and tangents to detect ellipse centre

To find the parameters a, b and c, the ellipse is firstly translated to be centered on the origin. Then the general equation (2) becomes:

$$ax^2 + 2bxy + cy^2 = 1 \quad (3)$$

Finally, three pixels of the ellipse are sufficient to resolve this equation:

$$\begin{bmatrix} x_1^2 & 2x_1y_1 & y_1^2 \\ x_2^2 & 2x_2y_2 & y_2^2 \\ x_3^2 & 2x_3y_3 & y_3^2 \end{bmatrix} \begin{bmatrix} a \\ b \\ c \end{bmatrix} = \begin{bmatrix} 1 \\ 1 \\ 1 \end{bmatrix} \quad (4)$$

### b- Features extraction

As in many computer vision-related problems, feature extraction is an important operation for the road signs detection and recognition.

The detected blobs are resized into 32 X 32 binary images. The haralick features and the pseudo Zernike moments are the extracted features from these images. However, a major problem associated with computer vision is the so-called curse of dimensionality [20]. There are, in fact, 14 parameters of Haralick. For this reason, we use PCA to know the most representative features to keep them, and discard the other elements of the vector. Haralick's texture features [21] are based on the gray-level co-occurrence matrix G:

The features used here to perform sign detection are:

- Homogeneity:

$$\text{hom} = \sum_i \sum_j p(i,j)^2 \quad (5)$$

With: Element p(i,j) is generated by counting the number of times a pixel with value i is adjacent to a pixel with value j.

- Correlation:

$$corr = \frac{[\sum_i \sum_j (ij)p(i,j) - \mu_x\mu_y]}{\sigma_x\sigma_y} \quad (6)$$

Where $\mu_x, \mu_y, \sigma_x$ and $\sigma_y$ are the means and standard deviations of $p_x$ and $p_y$, the partial probability density functions.

- Variation:

$$var = \sum_i \sum_j (i-\mu)^2 p(i,j) \quad (7)$$

- Difference variance:

$$DiffVar = \sum_{i=0}^{N_g} i^2 p_{x-y}(i) \qquad (8)$$

Where $p_{x-y}$ is the probability of co-occurrence matrix coordinates (x – y).

- Zernike moments [22]: $Z_{00}$ and $Z_{10}$.

**c- SVM classifier**

In supervised learning, it has been shown that discriminative classifiers have better performance than generative classifiers [23]. The theory of the support vector machine (SVM) [24] is very used in the binary class classification problem as well as in objects recognition and detection.

SVM classifier deals with two-category classification problems [23]. Given a training set $\{(x_i, y_i), i=1, …, n, x_i \in R^d, y_i \in \{+1, -1\}\}$, where $x_i$ is the feature vector, $y_i$ is the label. The aim of SVM is to maximize the margin between two categories besides distinguishing them.

## III- Results and discussion

### a- System description

The accuracy rate and the performance of the proposed approach are measured by applying it on a set of real data collected on Moroccan roads. This road sign detection system is implemented using C++ code and OpenCV2.2 library. The performance evaluation of our system is based on the recall and precision values, which are defined as follows:

$$recall = \frac{true\ positive\ detected\ (TP)}{total\ true\ positves} \times 100\% \qquad (9)$$

$$precision = \frac{true\ positive\ detected\ (TP)}{all\ detections} \times 100\% \qquad (10)$$

### b- Discussion

In order to check the improvements on the total accuracy and the measurement's precision, a comparative analysis has been done. We present a comparison of the performance of three methods of road signs detection:

- RHT
- Adaboost with pseudo Haar wavelet
- RHT + Texture

We tested our algorithm on a database of images acquired by a camera in urban areas. For each image, the number of true and false detection of road signs is counted. We find, after an analysis of the results, that more than 95% of the signs are correctly detected using our approach (Figure6). In addition, the false detection rate is decreased by more than 40% (Figure5) compared to the other two methods (RHT and Adaboost). This shows that our method has increased performances for detecting signs in an image.

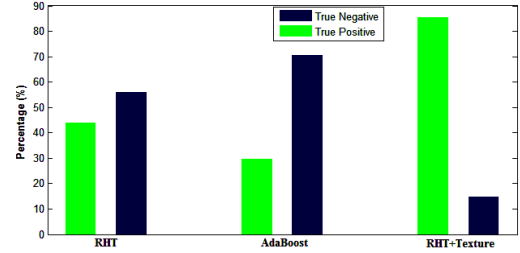

Figure5. The true positive and true negative accuracy

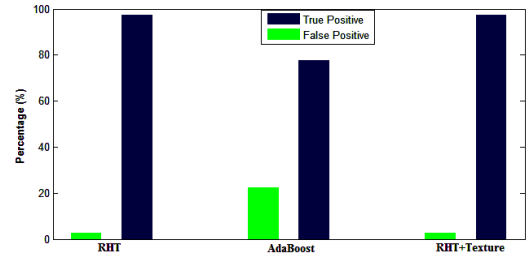

Figure6. The true positive and false positive accuracy

To show the robustness of our proposed solution we used precision-Recall curve shown in Figure7 where we can see that the Haar method gives the lowest performance among other methods of detecting signs in an image. Using the Hough transform shows better performance compared to the method of Haar. However, the proposed method (Hough + Texture features) shows the best performance among all others.

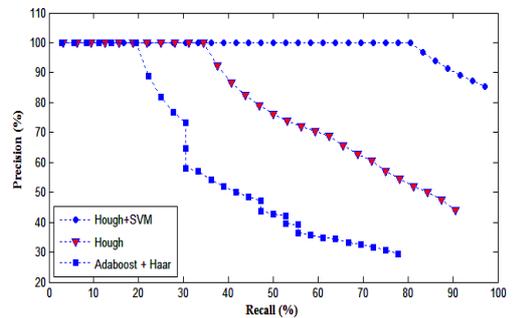

Figure7. Comparison between three methods in road signs detection

## IV- Conclusion

In this work, we improved the detection stage which represents an essential step in road signs identification. In addition, the quality of the detection affects the quality of recognition. Our technique decreases the number of false detection in an image and therefore improves the performance of our system by combining local and global features of the image. We used the method RHT, representing the locale method, as forms detector (octagonal and circular), and the combination of two global parameters represented by Zernike moments and Haralick features. The experimental results, as shown in Figure7 and Figure8, prove the robustness and effectiveness of the proposed approach.

**Acknowledgment**

The research is supported By Volubilis Project 2697WA (PHC MA/12/279).

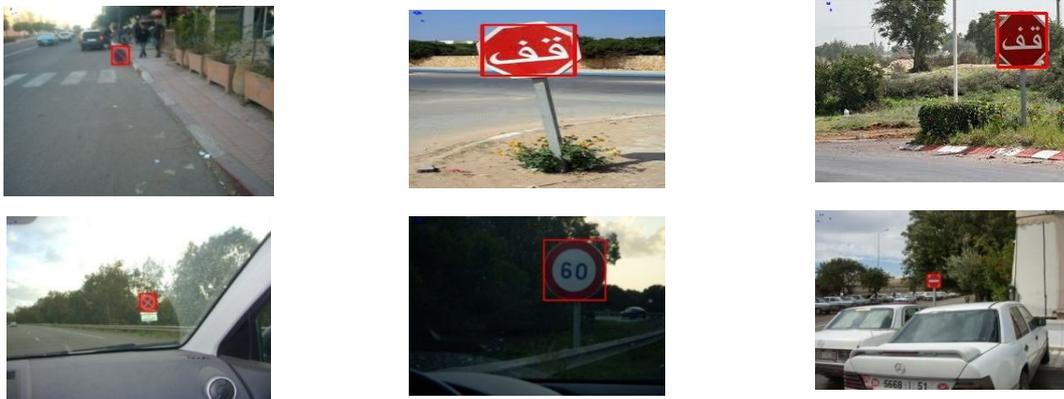

Figure8. Result of detection using our approach